\documentclass[preprint,12pt]{elsarticle}

\usepackage{amssymb}
\usepackage{amsthm}

\usepackage{amsmath,amsfonts}
\usepackage{algorithmic}
\usepackage{algorithm}
\usepackage{array}
\usepackage[caption=false,font=normalsize,labelfont=sf,textfont=sf]{subfig}
\usepackage{textcomp}
\usepackage{stfloats}
\usepackage{url}
\usepackage{verbatim}
\usepackage{graphicx}
\usepackage{cite}
\usepackage{xspace}

\newtheorem{theorem}{Theorem}
\newtheorem{definition}{Definition}
\newtheorem{proposition}{Proposition}
\newtheorem{corollary}{Corollary}
\newtheorem{lemma}{Lemma}
\newtheorem{example}{Example}

\newcommand{\bdfn}{\begin{definition}}
\newcommand{\edfn}{\end{definition}}
\newcommand{\bthm}{\begin{theorem}}
\newcommand{\ethm}{\end{theorem}}
\newcommand{\bprop}{\begin{proposition}}
\newcommand{\eprop}{\end{proposition}}
\newcommand{\blem}{\begin{lemma}}
\newcommand{\elem}{\end{lemma}}
\newcommand{\bcor}{\begin{corollary}}
\newcommand{\ecor}{\end{corollary}}
\newcommand{\bex}{\begin{example}}
\newcommand{\eex}{\end{example}}
\newcommand{\bobs}{\begin{observation}}
\newcommand{\eobs}{\end{observation}}

\newcommand{\framework}{Diversity Reduction Framework\xspace}
\newcommand{\fram}{DRF\xspace}

\journal{Artificial Intelligence}

\begin{document}

\begin{frontmatter}



\title{Reducing Diversity to Generate Hierarchical Archetypes}

\author[label1]{Alfredo Ibias\corref{cor1}}
\ead{alfredo@avatarcognition.com}
\author[label1]{Hector Antona}
\ead{hector@avatarcognition.com}
\author[label1]{Guillem Ramirez-Miranda}
\ead{guillem@avatarcognition.com}
\author[label1]{Enric Guinovart}
\ead{enric@avatarcognition.com}
\author[label2]{Eduard Alarcon}
\ead{eduard.alarcon@upc.ed}
\affiliation[label1]{organization={Avatar Cognition},
            city={Barcelona},
            country={Spain}}

\affiliation[label2]{organization={Universitat Politècnica de Catalunya - BarcelonaTech},
            city={Barcelona},
            country={Spain}}
\cortext[cor1]{Corresponding author}

\begin{abstract}
The Artificial Intelligence field seldom address the development of a fundamental building piece: a framework, methodology or algorithm to automatically build hierarchies of abstractions. This is a key requirement in order to build intelligent behaviour, as recent neuroscience studies clearly expose. In this paper we present a primitive-based framework to automatically generate hierarchies of constructive archetypes, as a theory of how to generate hierarchies of abstractions. We assume the existence of a primitive with very specific characteristics, and we develop our framework over it. We prove the effectiveness of our framework through mathematical definitions and proofs. Finally, we give a few insights about potential uses of our framework and the expected results.
\end{abstract}

\begin{highlights}
\item AI needs a framework for building hierarchies of abstractions.
\item We propose a framework to generate hierarchies of abstractions.
\item We assume a primitive with specific characteristics.
\item We outline a theory and prove some of its properties.
\item We show potential uses of the framework.
\end{highlights}

\begin{keyword}
Abstraction Algorithms \sep Diversity Reduction \sep Archetype Generation \sep Hierarchical Representations.
\end{keyword}

\end{frontmatter}

\section{Introduction}
Novel theories of the brain claim that our brains predict the world~\citep{hb04}. They do it by modelling the world in some form of abstraction, and then simulating what would happen. However, to model the world, first they need to extract, from a myriad of real world perceptions of, for example, an object (that is, a set of expressions of that object), the essence of such object in the form of a model or \emph{representation}. These representations are what in the Artificial Intelligence field has been termed \emph{an abstraction}, and what in this paper we aim to capture in the form of \emph{archetypes}. An archetype would be an object able to express the same behaviour than the abstraction it represents when manipulated, and it can be simple (with just one component) or composed (being the product of multiple individual components).

Coming back to how our brain works, it is important to know that we not only build standalone abstractions of objects, but instead we generate abstractions in a constructive and hierarchical way~\citep{fve91}. For instance, four wooden sticks and a wooden board make a wooden table, and the union of tables, chairs and other objects makes the concept of furniture. This kind of hierarchical thinking is known to be a major player in how cognition and intelligent behaviour develops~\citep{rb99}. Moreover, our own brain is known to be divided into a processing hierarchy, where higher levels of the cortex deal with more abstract concepts. For instance, a monkey's brain has less cortex levels and thus has more difficulties developing intelligent behaviours~\citep{fve91}.

This functioning arises a fundamental task: to develop frameworks to automatically build hierarchies of constructive abstractions. However, since the beginning of the Artificial Intelligence field, very few proposals have been focused on developing such frameworks. That has not impeded the development of a plethora of methods and algorithms that can simulate to work with such hierarchies, although none of them has been proven to actually build them. The most successful proposal, Deep Neural Networks~\citep{Aggarwal18}, is able to recognise thousands of different objects if trained properly~\citep{ksh17}, and even clusters them together in a latent space~\citep{Laine18}. However, no hierarchy of constructive abstractions is built and no cognition arises from such clustering.

Up to date, most of the developed algorithms have a huge dependency on a very basic assumption: they are optimising. This implies that the input and output spaces are assumed to be infinite mathematical spaces and the task becomes mapping from one to the other. However, as novel theories of the brain~\citep{hb04} (based on neuroscience research results) clearly expose, that is not how our brains work. In fact, our brain works in a constructive way, were abstractions are built based on the observed world. And then the brain organises such abstractions in a hierarchy, building higher level abstractions the higher in the hierarchy a concept is dealt with~\citep{rb99}. Thus, any proposal of artificial general intelligence algorithm should include a framework to automatically build such hierarchies of abstractions.

This change of perspective, from the mapping between mathematical fields to the constructive building of representations, has a major impact in how to approach the development of any artificial intelligence algorithm. In this scenario, we will no longer be able to work with input and output spaces, but instead we will need to work with input and output sets. This is derived from the fact that, in a constructive approach, whatever has not been seen during learning (and is not ``similar'' enough to previously seen samples) is not part of the input set and thus it does not have an internal representation. This implies that such new samples can not be recognised until learning resumes.

In this paper we present our proposal of a framework to automatically generate hierarchies of constructive abstractions through the use of archetypes. Such framework is based on the assumption that a primitive exists that builds constructive archetypes of its inputs. As our framework is a high level mechanism, we do not delve into the specifics of how such archetypes are generated, but instead we set some limits and characteristics that the primitive should fulfil for this framework to work as intended. Afterwards, once we have a description of the primitive, we prove how our framework builds a hierarchy of archetypes starting from the raw inputs. We prove our claims using mathematical proofs, and for that we present some fundamental definitions. Finally, we also present some use cases of our framework, showing its potential in the form of examples of architectures and expected outcomes. We also discuss some limitations of our framework, with special focus on the limitations of a pure theoretical work.

The rest of the paper is organised as follows. In Section~\ref{sec:work} we briefly present the related work of our framework. In Section~\ref{sec:fram} we present our framework. In Section~\ref{sec:proofs} we present the mathematical assumptions and proofs that support our work. In Section~\ref{sec:cases} we present some case scenarios to show the potential of our framework. In Section~\ref{sec:disc} we discuss some limitations of our framework. And finally, in Section~\ref{sec:conc} we present the conclusions of our work.

\section{Related Work}\label{sec:work}
The use of archetypes is not something new in the Artificial Intelligence field. Even less in science in general. In fact, automatic tools to generate archetypes have been used in fields as diverse as environmental simulations~\citep{gzcs22}, energy modelling~\citep{khp18} and video-game modelling~\citep{mtfg22}. But what all these fields have in common is that they fail to build hierarchies of those archetypes.

The use of hierarchies is nothing new neither. In fact, there is a huge field for generating hierarchies of neural networks~\citep{jdh98,zlhh17,sgcg+21}, and another huge field that tries to find hierarchies of representations in neural networks~\citep{ajylr18, cblz+20}. Also, there are some works in the reinforcement learning field that use hierarchies, with approaches like using a hierarchy for temporal abstractions~\citep{gp15,sss16,tgzmm17}. However, none of these proposals build constructive abstractions, and in the best cases the abstractions are pre-defined by the user.

Finally, the closest to a proposal of hierarchies of abstractions that we are aware of is the Rasmussen Hierarchy abstraction~\citep{Rasmussen86}, more recently called abstraction-decomposition space~\citep{vicente99}. It is a method to perform work domain analysis, a task focused on analysing and modelling process control systems. However, this is not general and thus is not close to what we aim to achieve in this paper.

Thus, to the best of our knowledge, there are no other proposals of a framework, methodology or algorithm to build hierarchies of constructive abstractions through archetypes.

\section{The \framework}\label{sec:fram}
Our proposal is a framework we called \emph{\framework} (\fram) that is based on a primitive. Thus, such primitive has a huge impact on the performance and capabilities of our algorithm. However, in this paper we do not present an example of such a primitive, but instead a series of characteristics that such a primitive has to have for the framework to work as intended. To be precise, we assume that the primitive over which the framework is built has the following characteristics:
\begin{itemize}
    \item Constructive approach: the primitive should not assume the existence of an input space, but instead take an input set and construct archetypes that will conform its output set.
    \item Same input shape than output shape: as the output set of the primitive will become the input set of another instance of the primitive, it is mandatory that their input and output sets are of the same shape.
    \item Reduction of diversity: the primitive should be able to produce a smaller output set than its input set. This way the primitive will be reducing the diversity present in the input set.
    \item Projections: the primitive should be able to generate projections of its archetypes, in order to transform any archetype into an input value.
\end{itemize}

 An example of primitive is displayed in Figure~\ref{fig:primi}, that receives an input and produces an output and a projection. We define a primitive $P$ as a process able to receive a finite input set $I$ and produce an output set $O$ with less elements than $I$. However, $O$ can not be any output set, it has to be an output set for which it exists a ``reverse'' function $f$ that, given an element $o\in O$, it produces an element $i\in I$ such that it is one of the elements that $P$ maps to $o$. That is, in a more formal way, if $f(o) = i$ then $P(i) = o$, thus $P(f(o)) = o$. However, it is not always true that $f(P(i)) = i$, as it is not mandatory that the recovered input is the same, but only that it is one of the inputs mapped to $o$. This output set $O$ is what we called a \emph{latent set} of $I$, because it contains enough information to represent the input set $I$ with less elements, and thus we consider its elements to be archetypes of the elements of $I$. Additionally, $f$ will be the projection function of the primitive. A more formal definition of these components is presented in the next section.

Reducing diversity is the fundamental task of the primitive, as it is the base assumption over which most of the framework is built (as will be made clear in Theorem~\ref{thm:diversity}). We define diversity as the number of different elements in a set, in a similar fashion than previous work~\citep{bc22}. This property ensures that we apply the principle of ``to comprehension through compression''~\citep{Wagensberg14}, where we compress the input set into a smaller output set (without repeating elements inside them). Once we have a primitive with those characteristics, our \framework organises multiple instances of such primitive into a hierarchical structure. The final goal is to generate archetypes over archetypes. In that sense, each instantiated primitive has a different input set and a different hierarchy level.

The lower level input sets are conformed by the input signals, and the primitives that receive these input sets are the lowest level primitives. These primitives transform those input sets into archetypal output sets, thanks to their required reduction of diversity.
These archetypes will be of the same shape of the input they represent, and will conform part of the input set of a higher level primitive.

A higher level primitive takes as input set the composition of the output sets of multiple lower level primitives. Thus, this higher level primitives apply the same archetype principle but over an already archetypal domain with lower diversity, allowing for the further reduction of diversity and thus the generation of higher-order archetypes. Reproducing this setup along multiple levels is how we reduce the diversity of the input set.

This simple organisation needs an extra element: the method by which the framework composes signals from the lower levels. We propose two options for such composition: averaging the lower level outputs, or concatenating them.

With the first option we will be building compositions of archetypes that would improve the discriminability of the primitive. This type of composition is more suited for identifying samples, like an image of a table versus an image of a chair. A structure that uses the averaging composition is what we call a \emph{discriminatory pyramid}, and an example of it is displayed in Figure~\ref{fig:pyramid}. A fundamental property of any discriminatory pyramid is that it builds a latent set of its unique input set as long as the primitive does so, thus the final output set is a smaller latent set of the global input set. A more formal definition of this structure is presented in the next section.

If we choose the second option we will be building compositions of archetypes that would focus on the association between them. This type of composition is more suited for associating patterns, like an image of a table and its sound when moved. A structure that uses the concatenation composition is what we call an \emph{associative layer}, and an example of it is displayed in Figure~\ref{fig:layer}. A fundamental property of any associative layer is that it builds a latent set of the associations between its different input sets, and if its input sets are latent sets then the final output set is a latent set of the associations between those input latent sets, and thus between the global input sets. A more formal definition of this structure is presented in the next section.

One extra requirement of our proposal is that any archetype can be materialised into an actual value of the input set, as we need to be able to materialise the archetypes in order to interpret them. Thus, a kind of double way should be built, where in one pass the inputs are being archetyped and in another pass the archetypes are being materialised into input values. To that end, the requirement of our primitive that the input and output shapes should be the same is primordial, as that allows for working with the same shape in any direction. It is also fundamental the requirement that the primitive can build projections, as those will be the materialisation of the archetypes that will be used in the downward pass.

An illustrative example of the intuition behind how this framework would work will be the furniture example: let us start by creating a different primitive for each possible material of table, one for wooden tables, other for metal tables, another more for plastic tables, etc... These primitives will take in all the diversity of possible tables of their material and archetype them in few archetypes, simplifying in that sense all the possible tables of one material into an archetype of table of that material. We can make a similar deployment for chairs, sofas, etc... Now, going one level up, we will deploy a primitive that receives as input all the outputs of the material table primitives. Thus, this table primitive will compose all the archetypes of the different table materials, and build few archetypes of tables. For example, it can build the archetypes of circular tables, square tables, etc... independently of their material. Then, we have an archetype of tables. Similar deployment can be done for chairs, sofas, etc... Finally, in the final level, we can deploy a primitive for furniture, that will take as input the different furniture archetypes generated in the previous level and build archetypes of furniture as a concept.

This example gives the basic intuition about how a discriminatory pyramid work, building archetypes of each time more complex concepts. To give the intuition of how an associative layer works we need a more complex example, were we have inputs with different typology that have an association between them. A basic example of this will be the pattern matching example. Let us have a dataset with samples and associated labels, then we will build an associative layer that receives as inputs in one side the sample and in the other side the label. It will then build archetypes of their relationships. This way, the primitive will detect patterns via building associative archetypes of pairs of sample and label.

In this last example, the requirement of our proposal of being able to materialise any archetype into an actual value is very useful. It allows us to provide the associative layer with one sample, and use the archetypes generated by the primitive of the associative layer to find the label associated with such sample. This would allow to classify previously unseen samples using our learned archetypes and their association with labels.

Finally, a mixed structure, with lower level primitives that use the average composition and higher level primitives that use the concatenation composition, could potentially develop higher level archetypes of very good quality. In that sense, for each input type, we would have a discriminatory pyramid taking such input, transforming it into an archetype, and averaging their archetypes in a pyramidal structure until the highest level primitive has the most defined archetype able to identify the sample. Then, once we have one of these structures for each input type, we can start associating those archetypal inputs using associative layers that receive as input the outputs of the discriminatory pyramids. The discriminatory pyramids have properly identified the input and the associative layers only find the relationships between different inputs, until the highest level primitive finds the total relationship between all the inputs.

A good example for this last case would be the association between the sounds of tables and images of tables. Having a discriminatory pyramid identifying the sounds and splitting between the sounds of metal tables and those of wooden tables or plastic tables, and then having another discriminatory pyramid identifying the images of tables and splitting between the images of metal tables and those of wooden tables or plastic tables, allows us to have an associative layer generating archetypes of the association between the sound and image of different tables based on their material. As the inputs would always come synchronised (an image of a wooden table will always come with a sound of a wooden table), then the resulting structure is able to identify the material of a table based on either its sound or its image. Moreover, it is able to recover, from an image of a table, a sample of the sound it could produce. Note here that, for this example, there would be a need for the transformation of both sound and image to a same datatype (like an Sparse Distributed Representation) in order to be able to process them with the same kind of primitive. However, as that is an implementation concern, in the rest of the paper we will assume that those details are being taken care of.

\section{Proofs}\label{sec:proofs}
In this section we will present some definitions, theorems and proofs that will prove that our framework does what we claim. First, we will delve into the properties and efectiveness of the primitive. Later, we will focus on the scalability component of our framework in two steps: first focusing on the discriminatory pyramids, and later in the associative layers.

\subsection{The primitive and its properties}
Let us start by proving some theorems about the primitive that are derived from its properties. First, we need to define a preliminary building block: a process.
\bdfn
A process $\mathcal{P}$ is any collection of transformations and operations that transform elements from an input set $I$ to elements of an output set $O$.

A process $\mathcal{P}$ can be parameterised, and thus generate different instances of it with different values of its parameters.
\edfn
This definition of process is the building block over which we will build our definition of primitive.

Now, we need to prove that any process is equivalent to a mathematical surjective function as long as its output set is equal or smaller than its input set.
\bthm\label{thm:map}
Given a process $\mathcal{P}$, and given an input set $I$, if $\mathcal{P}(I)$ produces an output set $O$, with $|O| \leq |I|$, then $\mathcal{P}$ is equivalent to a surjective function mapping $I \to O$.
\ethm
\begin{proof}
    If the input set $I$ has size $|I| = n$, and the process $\mathcal{P}$ processes $I$ and produces an output set $O$ with size $|O| \leq n$, then there is an association between the sets $I$ and $O$ in the form of pairs composed of an input $i\in I$ and an output $o\in O$.
    
    There will be as many pairs as inputs, and if $|O| = n$ no output will appear in more than one pair. Otherwise, by the pigeonhole principle, there will be at least one output $o$ appearing in more than one pair.

    If we take these pairs, we can build a surjective function that maps, for each pair, its input to its output. Thus, there exists a surjective function that maps $I \to O$ in the same fashion than $\mathcal{P}$. And $\mathcal{P}$ will be equivalent to such function.
\end{proof}
This result allows us to reason about processes as mathematical functions, what will be necessary in future proofs.

We also need to prove that any process that produces an smaller output set is reducing the diversity that comes from its input set.
\bthm\label{thm:diversity}
Given a process $\mathcal{P}$, and given an input set $I$, if $\mathcal{P}(I)$ produces an output set $O$, with $|O| < |I|$, then $\mathcal{P}$ is a function that reduces diversity.
\ethm
\begin{proof}
    If the input set $I$ has size $|I| = n$, and the process $\mathcal{P}$ processes $I$ and produces an output set $O$ with size $|O| < n$, then, by the pigeonhole principle, there exists at least two inputs $i_1, i_2$ and an output $o$ such that $\mathcal{P}(i_1) = o$ and $\mathcal{P}(i_2) = o$. Thus, the process $\mathcal{P}$ has reduced the cardinality of the set $I$ by transforming it into the set $O$.
    
    As both $I$ and $O$ are sets, there are no repeated elements in them, thus, there has been a reduction in diversity through the reduction in size of the original set.
\end{proof}
This result is fundamental, as proves the minimum requirements of any process to reduce diversity. Here it is important to remember that, as we are working with mathematical sets, those do not contain repeated elements. This fact ensures that the number of different elements is reduced through the application of the primitive.

Once we have a process that reduces the diversity of the input set, we need to prove that such process is reducing the diversity in a meaningful way. This implies that the output set should be a latent set of the input set. First, we need to define a latent set.
\bdfn\label{def:latent}
A latent set $L$ of a bigger set $S$ is a set of elements with $|L| < |S|$, such that there exists a surjective function $f:S \to L$ that maps each element $s \in S$ to an unique element $l \in L$.

Additionally, there exists a reverse injective function $g: L \to S$ that maps each element $l\in L$ to an unique element of $s\in S$, and such element $s$ is one to the elements of $S$ that are mapped to $l$ via $f$.

During the rest of the paper, abusing notation, we will mark $g$ as $f^{-1}$, although we are aware that it is not the inverse function of $f$ .
\edfn
This definition of a latent set will allow us to differentiate between the generation of archetypes and the generation of random mappings.

Having a latent set will allow us to recover elements of the input set from their archetypes in the output set. This in fact would imply that the process is properly doing an archetype, and that is the reason why one of the requirements of the primitive was to be able to produce projections.
\bthm
Given an input set $I$, and given a process $\mathcal{P}$ that produces an output set $O$ such that $|O| < |I|$, if there exists an inverse injective function $\mathcal{P}^{-1}$ such that $\mathcal{P}^{-1}: O \to I$ is able to produce a valid element of the input set $i\in I$ for each element of the output set $o\in O$ in a way that $\mathcal{P}(\mathcal{P}^{-1}(o)) = o$, then the set $O$ is a latent set of the input set $I$.
\ethm
\begin{proof}
    Let us assume $O$ is not a latent set of the input set $I$. Then, by Definition~\ref{def:latent}, there will be no injective function $f:O\to I$ such that $f$ maps each element $o\in O$ to an unique element of $i\in I$, and such element $i$ is one of the elements that are mapped to $o$ via $\mathcal{P}$.
    
    Now, let us assume that there exists an injective function $\mathcal{P}^{-1}: O \to I$ such that $\mathcal{P}(\mathcal{P}^{-1}(o)) = o$. As $\mathcal{P}^{-1}$ is an injective function from $O$ to $I$, there exists an element $i \in I$ such that $\mathcal{P}^{-1}(o) = i$, thus $\mathcal{P}(\mathcal{P}^{-1}(o)) = \mathcal{P}(i) = o$. However, as the process $\mathcal{P}$ is equivalent to a surjective function by Theorem~\ref{thm:map}, $\mathcal{P}(i) = o$, and $i\in I$, then $i$ is one to the elements of $I$ that are mapped to $o$ via $\mathcal{P}$.
    
    Thus, $\mathcal{P}^{-1}$ is an injective function $O\to I$ such that it maps each element $o\in O$ to an unique element of $i\in I$, and such element $i$ is one of the elements that are mapped to $o$ via $\mathcal{P}$. Thus, $O$ is a latent set of the input set $I$.
\end{proof}
This result proves that the kind of process that we require is properly building archetypes of the inputs it receives. This is a fundamental result that is key for our framework. Later, when we compose multiple primitives in the fashion we described in Section~\ref{sec:fram}, this result will be a fundamental building block. Thus, we will define the concept of primitive as a process whose output set is a latent set of the input set.
\begin{figure}[t]
    \centering
    \includegraphics[width=0.7\columnwidth]{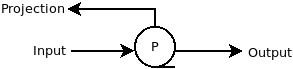}
    \caption{Basic structure of a primitive.}
    \label{fig:primi}
\end{figure}
\bdfn\label{def:prim}
A primitive $P$ will be a process that transforms an input set $I$ into a latent set $O$ of the input set.

That is, $P$ is a process from $I$ to $O$, with $|O| < |I|$, and such that there exist an inverse injective function $P^{-1}: O \to I$ that is able to produce a valid element of the input set $i\in I$ for each element of the output set $o\in O$ in a way that $P(P^{-1}(o)) = o$. This function $P^{-1}$ is the projection function of the primitive $P$.
\edfn
An example of primitive is displayed in Figure~\ref{fig:primi}, that receives an input and produces an output and a projection.

An important note about the definition of primitive is that a trivial primitive will be one that merges all inputs into an unique output. Although this sounds like a problem, it is in fact a required property, because we need to be able to represent concepts, that is, archetypes that represent all samples in a single element. For example, the concept of \emph{numbers} merges all samples of numbers into one archetype: \emph{numbers}. Then, when someone ask us for a number, we provide one sample of such archetype, thus producing a projection. However, although we need to allow for this extreme case, it is far from ideal for a practical application, as it would impede the composition of such primitive with other instances of itself due to its output set being already only one sample, what makes it impossible of further reduction.

\subsection{Discriminatory Pyramids}
Let us now analyse the discrimination power of the discriminatory pyramids. First, we need to define what is a discriminatory pyramid.
\bdfn
Given a primitive $P$, a \emph{discriminatory pyramid} $DP$ is a hierarchical structure composed of multiple instances of the primitive $P$. A pyramid has $n$ levels, and in each level it has $2^l$ primitives, with $l$ being the level number starting from the top with the level $0$.

The $2^n$ primitives of level $n$ receive all of them the same input, an element $i\in I$, and the final output of the pyramid is the output of the primitive of the level $0$, that is an element $o\in O$. Thus, $I$ is the input set of the pyramid and $O$ is its output set.

Finally, between each level there is an average of the lower level outputs. These averaged values will be the inputs of the higher level.
\edfn
\begin{figure}[t]
    \centering
    \includegraphics[width=0.7\columnwidth]{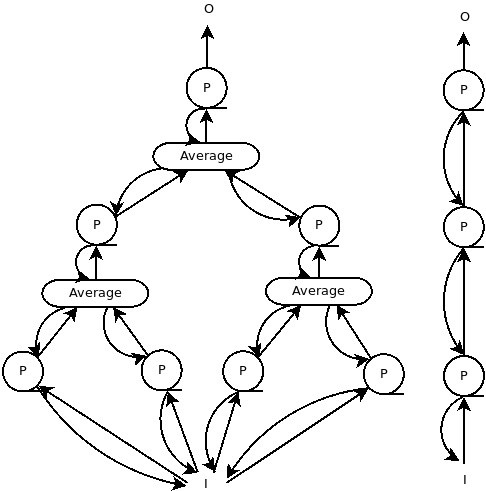}
    \caption{An example of discriminatory pyramid (left) and of its base case called discriminatory column (right).}
    \label{fig:pyramid}
\end{figure}
An example of discriminatory pyramid is displayed in Figure~\ref{fig:pyramid} (left). With this definition, we can start analysing its properties. First, let us show that the averaging of two latent sets is still a latent set.
\bthm\label{thm:latent_avg}
Given a primitive $P$, an input set $I$, two latent sets of that input set $O_1$ and $O_2$ produced by two different instances of $P$, and a function $f:(O_1\times O_2) \to O$ that computes the average of the inputs, then $O$ is a latent set of $I$.
\ethm
\begin{proof}
    Let us assume $O$ is not a latent set of $I$. Then, by Definition~\ref{def:latent}, there will be no injective function $f:O\to I$ such that $f$ maps each element $o\in O$ to an unique element of $i\in I$ and such element $i$ is one of the elements that are mapped to $o$ via $P$.
    
    Now, each element $o \in O$ is an average of other two elements $o_1 \in O_1$ and $o_2 \in O_2$. Thus $o = \frac{o_1 + o_2}{2}$.
    
    We also have that both $o_1$ and $o_2$ have been produced from the same input $i \in I$ through two instances $P_1$ and $P_2$ of the primitive $P$. Thus $o = \frac{P_1(i) + P_2(i)}{2}$.

    If we assume, without loss of generality, that $|O_1| \geq |O_2|$, then we have that $|O| = |O_1|$. Thus $|O| < |I|$.
    
    Now, we know of the existence of two injective functions $P_1^{-1}$ and $P_2^{-1}$ such that $P_1^{-1}(o_1) = i$ and $P_2^{-1}(o_2) = i$ respectively. Thus, we can build a function $f$ such that
    \begin{align*}
        f_{o_2}(o)&=P_1^{-1}(2\cdot o - o_2) \\
         &=P_1^{-1}(2\cdot o - P_2(i)) \\
         &=P_1^{-1}(2\cdot \frac{P_1(i) + P_2(i)}{2} - P_2(i)) \\
         &=P_1^{-1}(P_1(i)) \\
         &=i
    \end{align*}
    for all $o \in O$.

    Finally, $f$ is an injective function because $|O| = |O_1|$ and $P_1^{-1}$ is already an injective function.

    Thus, there exists an injective function $f:O\to I$ such that $f$ maps each element $o\in O$ to an unique element of $i\in I$ and such element $i$ is one of the elements that are mapped to $o$ via $P$.
\end{proof}
With this result we can claim that averaging latent sets still produces a latent set. However, unlike other proofs that will come later, this proof does not provide us with an useful projection function for implementation, as it will depend on the specifics of the implemented primitive. This will make the implementation of the discriminatory pyramids harder, but not impossible.

Now, let us prove that the latent set of a latent set is still a latent set of the original set.
\bthm\label{thm:latent_composition}
Given a primitive $P$, an input set $I$, a latent set of that input set $O_1$ produced by $P$, and a latent set $O_2$ of the latent set $O_1$ produced by $P$, then $O_2$ is a latent set of the input set $I$. 
\ethm
\begin{proof}
    Let us assume $O_2$ is not a latent set of $I$. Then, by Definition~\ref{def:latent}, there will be no injective function $f:O_2\to I$ such that $f$ maps each element $o\in O_2$ to an unique element of $i\in I$ and such element $i$ is one of the elements that are mapped to $o$ via $P$.

    Now, we have that, by Definition~\ref{def:latent}, there exists functions $f_1$ and $f_2$ such that $f_1$ maps each element $o_1\in O_1$ to an unique element $i\in I$ with $P(i) = o_1$, and $f_2$ maps each element $o_2\in O_2$ to an unique element $o_1\in O_1$ with $P(o_1) = o_2$. Then, the composition of $f_2 \circ f_1$ is a function that maps each element $o_2\in O_2$ to an unique element of $i\in I$.
    
    Now, such element $i$ is mapped to $o_1\in O_1$ via $P$, that is, $P(i) = o_1$. And this element $o_1$ is mapped to $o_2\in O_2$ via $P$, that is $P(o_1) = o_2$. Thus, $i$ is one of the elements that are mapped to $o_2$ via $P$. Thus, $O_2$ is a latent set of $I$.
\end{proof}
This result proves the transitivity property of latent sets, what will be useful in following proofs.

Finally, let us show how the output set $O$ of a discriminatory pyramid is still a latent set of its input set $I$.
\bthm
Given a primitive $P$ and given a discriminatory pyramid $DP$ composed of $n$ levels of instances of the primitive $P$, with an input set $I$ and an output set $O$, then $O$ is a latent set of $I$.
\ethm
\begin{proof}
    Let us assume $O$ is not a latent set of the input set $I$. Then, by Definition~\ref{def:latent}, there will be no injective function $f:O\to I$ such that $f$ maps each element $o\in O$ to an unique element of $i\in I$, and such element $i$ is one of the elements that are mapped to $o$ via $P$.
    
    Now, by Definition~\ref{def:prim}, each instance of the primitive $P$ takes its input set and transforms it into a latent set of it. Thus, any primitive in the $n$th level will transform the input set $I$ into a latent set of it $O_{n-1_i}$.
    
    To conform the input set of the primitives of level $n-1$, an average of the outputs of each pair of primitives of level $n$ is computed. This process transforms two output sets that are latent sets of $I$ into an unique output set $O_{n-1}$ that is a latent set of the input set $I$ due to Theorem~\ref{thm:latent_avg}.

    At level $0 < m < n$, the primitives receive a latent set $O_m$ of the input set $I$ and transform them into a latent set $O_{m-1_i}$ of the set $O_m$, but by Theorem~\ref{thm:latent_composition} these new latent sets are latent sets of $I$. And the average function transform those latent sets $O_{m-1_i}$ into latent sets $O_{m-1}$ of the latent sets $O_{m-1_i}$, that by Theorem~\ref{thm:latent_composition} are latent sets of $I$.

    Finally, at level $m=0$, the last primitive receives a latent set $O_1$ of the input set $I$ and transform it into a latent set $O$ of the set $O_1$, that by Theorem~\ref{thm:latent_composition} is a latent set of $I$. Thus, $O$ is a latent set of $I$.
\end{proof}
This result proves that our discriminatory pyramids are still producing archetypes.

Now, let us prove that the archetypes produced by a discriminatory pyramid are more refined than the archetypes of a single primitive.
\bthm
Given a primitive $P$ and given a discriminatory pyramid $DP$ composed of $n\geq1$ levels of instances of the primitive $P$, both with an input set $I$, and $P$ producing an output set $O_P$ and $DP$ producing an output set $O_{DP}$, then $|I| > |O_P| > |O_{DP}|$
\ethm
\begin{proof}
    Let us start by considering an individual primitive $P$ with input set $I$, then $P$ produces an output set $O_P$ such that $|O_P| < |I|$.
    
    Now, let us consider level $n$ of $DP$. In this level, the inputs of each group of two primitives are the same, and both of them transform the input set $I$ into two output sets $O_{n-1_1}$ and $O_{n-1_2}$ with $|I| > |O_{n-1_1}|$ and $|I| > |O_{n-1_2}|$.
    
    Now, the input set of the average function is not all the possible combinations of values of $O_{n-1_1}$ and $O_{n-1_2}$, but instead the set of pairs $\{(o_1, o_2) | o_1\in O_{n-1_1}, o_2\in O_{n-1_2}, o_1 = P_{n_1}(i), o_2 = P_{n_2}(i)\}$. Having in account that the output set $O_{n-1}$ of the average function has as many values as its input set, we have that $|O_{n-1}| = max(|O_{n-1_1}|, |O_{n-1_2}|)$, and thus $|I| > |O_{n-1}| \approx |O_P|$.

    Now, for any level $0 < m < n$, we have that each group of two primitives have two different input sets $O_{m_1}$ and $O_{m_2}$, with $|I| > |O_{m_1}|$ and $|I| > |O_{m_2}|$. The two involved primitives transform them into two output sets $O_{m-1_1}$ and $O_{m-1_2}$ with $|O_{m_1}| > |O_{m-1_1}|$ and $|O_{m_2}| > |O_{m-1_2}|$. Now, the input set of the average function will be the set of pairs $\{(o_1, o_2) | o_1\in O_{m-1_1}, o_2\in O_{m-1_2}, o_1 = P_{m_1}(o_{m_1}), o_2 = P_{m_2}(o_{m_2})\}$. Having in account that the output set $O_{m-1}$ of the average function has as many values as its input set, we have that, as both input sets have a cardinality lower than $O_{m_{max}} = max(|O_{m_1}|, |O_{m_2}|)$, then $|I| > O_{m_{max}} > |O_{m-1}|$.

    Finally, at level $m=0$, the last primitive receives an input set $O_1$ with $|I| > |O_1|$ and produces an output set $O_{DP}$ with $|O_{DP}| < |O_1| < |I|$. Thus, $|I| > |O_P| > |O_{DP}|$.
\end{proof}
This is a fundamental result that proves that using discriminatory pyramids improves results over using a single primitive.

An interesting corollary of previous results is that, if the primitive is the same in each instance of it inside the discriminatory pyramid, then the pyramid is equivalent to a column. For that, first we need to define what is a discriminatory column.
\bdfn
Given a primitive $P$, a \emph{discriminatory column} $DC$ is a hierarchical structure composed of multiple instances of the primitive $P$. A pyramid has $n$ levels, and in each level it has $1$ primitive.

The primitive of level $n$ receives as input an element $i\in I$, and the final output of the pyramid is the output of the primitive of the level $0$, that is an element $o\in O$. Thus, $I$ is the input set of the pyramid and $O$ is its output set.

Finally, each level receives as inputs the lower level outputs.
\edfn
An example of discriminatory column is displayed in Figure~\ref{fig:pyramid} (right). Now, we can formulate the following corollary.
\bcor
Given a primitive $P$ and given a discriminatory pyramid $DP$ composed of $n$ levels of instances of the primitive $P$, if all the instances of $P$ produce the same output set $O$ given the same input set $I$, then $DP$ is equivalent to a discriminatory column $DC$ composed of $n$ levels of instances of the primitive $P$.
\ecor
\begin{proof}
    Let us start considering level $n$. In this level, all primitives receive the same input. Thus, their input set for all is $I$, and thus their output set is the same set $O_{n-1}$ for all of them. Then, the average function will have as input set the set $\{(o, o) | o \in O_{n-1}\}$, and as $\frac{o + o}{2} = o$ $\forall o\in O_{n-1}$, then the output set of the average function is still $O_{n-1}$. Thus, all the primitives of this level are equivalent to having a unique primitive in this level.

    Now, at levels $0 < m < n$ we have a similar situation. All primitives have the same input set $O_m$ and will produce the same output set $O_{m-1}$. The average function will be equivalent to the identity function again then. Thus, all the primitives of these levels will be equivalent to having a unique primitive at each level again.

    Finally, at level $0$, we already have a single primitive, so it is equivalent to having a single primitive in this level. Thus, $DP$ is equivalent to $DC$.
\end{proof}
This corollary is useful to save resources when the used primitive is not parameterised neither have any associated randomness in its process, and thus it always generates the same archetypes. It will also allow us to implement a discriminatory pyramid easier, as here the proof gives us a viable way to generate the global projection function.

\subsection{Associative Layers}
Finally, let us analyse the associative power of the associative layers. First, we need to define what is an associative layer.

\begin{figure}[t]
    \centering
    \includegraphics[width=0.5\columnwidth]{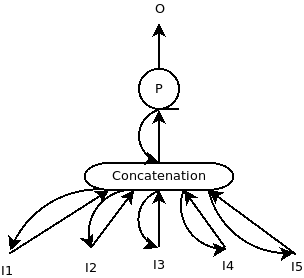}
    \caption{An example of associative layer.}
    \label{fig:layer}
\end{figure}
\bdfn
Given a primitive $P$, an \emph{associative layer} $AL$ is a structure composed of a concatenation function and an instance of the primitive $P$.

The concatenation function receives multiple inputs $i_1, \cdots, i_n$ from different input sets $I_1, \cdots, I_n$ and concatenate them into a single input $i\in I$, and $P$ process $i$ to produce an element $o\in O$. Thus, $I_1, \cdots, I_n$ is the input set of the associative layer and $O$ is its output set.
\edfn
An example of associative layer is displayed in Figure~\ref{fig:layer}.

Now, let us prove that a concatenation function generates a composition of input sets that keeps the association between correlated inputs.
\bthm\label{thm:corr}
Given correlated input sets $I_1, \cdots, I_n$ and a concatenation function such that it generates an output set $I$ that is a composition of the input sets. If there is no correlation between the inputs $i_1, \cdots, i_n$, then the output $i_1\cdots i_n$ is not present in $I$.  
\ethm
\begin{proof}
    Let us start by assuming that there is a value $i_1\cdots i_n \in I$ such that there is no correlation between at least two elements of $\{i_1, \cdots, i_n\}$. For example, without loss of generality, let us assume such two inputs are $i_j\in I_j$ and $i_k\in I_k$. Then, as there is no correlation between $i_j$ and $i_k$ but there is correlation between $I_j$ and $I_k$, then the inputs will not appear at the same time as inputs of the system. That implies that the concatenation layer will never receive an input such that $i_j$ and $i_k$ are part of such input. Thus, the concatenation layer will never produce the output $i_1\cdots i_j \cdots i_k \cdots i_n$, and thus such output will never be part of the output set $I$.
\end{proof}

It is also important to prove that, if there is no correlation between the inputs, then the concatenation function will produce all possible combinations of its inputs.
\bthm\label{thm:ind}
Given independent input sets $I_1, \cdots, I_n$ and a concatenation function such that it generates an output set $I$ that is a composition of the input sets. Then the output set $I$ is $\{i_1\cdots i_n| i_1\in I_1, \cdots, i_n\in I_n\}$.  
\ethm
\begin{proof}
    Let us start by assuming that there is a value $i_1\cdots i_n \notin I$ with $\{i_1 \in I_1, \cdots, i_n \in I_n\}$. This implies that there is at least two elements of $\{i_1, \cdots, i_n\}$ that will never appear together as inputs of the concatenation function. For example, without loss of generality, let us assume such two inputs are $i_j\in I_j$ and $i_k\in I_k$. Then, as there is no correlation between $I_j$ and $I_k$, there should be no correlation between $i_j$ and $i_k$. Thus, the inputs will eventually appear at the same time as inputs of the system. That implies that the concatenation layer will eventually receive an input such that $i_j$ and $i_k$ are part of it. Thus, the concatenation layer will eventually produce the output $i_1\cdots i_j \cdots i_k \cdots i_n$, and thus such output is part of the output set $I$.
\end{proof}
These two results are the cornerstone over which the associative power of the associative layer resides. The associations are performed at this level, while the primitive only generates archetypes of these associations.

Now, let us prove that the concatenation of latent sets is a reversible composition of latent sets.
\bthm\label{thm:latent_concat}
Given latent sets $I_1, \cdots, I_n$, and a bijective function $f:(I_1\times \cdots \times I_n) \to I$. Then, $I$ is a composition of latent sets that can be reverted.
\ethm
\begin{proof}
    First, let us state that, as $f$ is bijective, it exists a function $f^{-1}$ such that $f^{-1}(f(i)) = i$ $\forall i \in (I_1\times \cdots \times I_n)$. Thus, the function is revertible.
    
    Now, if we assume that $I$ is not a composition of latent sets, that means that either one of the input sets is not a latent set, or some of those input sets are combined in a way that makes them loose their latent set property. As all the input sets are latent sets, the only option is that some of the input sets are combined, but as $f$ is bijective, then this combination is not possible. Thus, the output set $I$ is a composition of the latent sets.
\end{proof}
With this result we prove that we only need a bijective function. Now, we just need to prove that the concatenation function is bijective.
\blem\label{lem:concat}
Given latent sets $I_1, \cdots, I_n$, and a concatenation function $f$, then $f$ is a bijective function $f:(I_1\times \cdots \times I_n) \to I$.
\elem
\begin{proof}
    Let us assume $f$ is not bijective, then $f$ is either not injective or not surjective.
    
    If $f$ is not surjective then there exists an input $i_1, \cdots, i_n \in (I_1\times \cdots \times I_n)$ such that, for at least two of its components it is impossible to recover the original inputs from the generated output. For example, without loss of generality, let us assume such two components are $i_j\in I_j$ and $i_k\in I_k$.
    
    Now, $f$ generates the output $i_1\cdots i_j \cdots i_k \cdots i_n$ when provided with the input $i_1, \cdots, i_j, \cdots, i_k, \cdots, i_n$. Thus, from the output we can recover the values of $i_j\in I_j$ and $i_k\in I_k$ with the function that given an index it gives back the values of such index. Thus, the original inputs can be recovered, what implies that $f$ is surjective.
    
    Finally, if $f$ is not an injective function, then there exists two inputs $i_1 = i_{1_1}, \cdots, i_{n_1} \in (I_1\times \cdots \times I_n), i_2 = i_{1_2}, \cdots, i_{n_2} \in (I_1\times \cdots \times I_n)$ such that $f(i_1) = f(i_2)$ and $i_1 \neq i_2$. For example, without loss of generality, let us assume that $i_1$ and $i_2$ differ in the indices $j$ and $k$, that is, $i_{j_1} \neq i_{j_2}$ and $i_{k_1} \neq i_{k_2}$.
    
    As $f$ is a concatenation function, $f(i_1) = o_1 = i_{1_1} \cdots i_{j_1} \cdots i_{k_1} \cdots i_{n_1}$ and $f(i_2) = o_2 = i_{1_2} \cdots i_{j_2} \cdots i_{k_2} \cdots i_{n_2}$. Thus, $o_1 \neq o_2$, and thus $f$ is a injective function.
    
    And an injective and surjective function is a bijective function.
\end{proof}
Thus, with these results we prove that we can recover the original latent sets from a composition of latent sets. This is fundamental to be able to recover the individual elements of a relationship.

Finally, let us prove that the primitive keeps the associations and generates a latent set of relationships.
\bthm\label{thm:al}
Given a primitive $P$, input sets $I_1, \cdots, I_n$ and a concatenation function that generates an output set $I$ that is a composition of those input sets. Then $P$ produces an output set $O$ that is a latent set of the relationships between $I_1, \cdots, I_n$.
\ethm
\begin{proof}
    As $P$ is a primitive, it generates a latent set of its input set. Thus, $O$ is a latent set of $I$. As $I$ is a composition of input sets, then $O$ is a latent set of the composition of input sets. Moreover, as $I$ keeps the relationships between the input sets due to Theorems~\ref{thm:corr} and~\ref{thm:ind}, then $O$ is a latent set of such relationships. Thus, $O$ is a latent set of the relationships between $I_1, \cdots, I_n$.
\end{proof}

Additionally, as a last proof, we want to prove that the use of a primitive after an associative layer generates a new latent set of associations.
\bcor
Given a primitive $P$ and an associative layer $AL$ that implements the same primitive $P$. Then if $I_1, \cdots, I_n$ are the input sets of $AL$, and $AL$ generates an output set $I$ after processing them, and $P$ processes $I$ and generates an output set $O$, then $O$ is a latent set of the relationships between $I_1, \cdots, I_n$.
\ecor
\begin{proof}
    As $AL$ produces a latent set of the relationships between $I_1, \cdots, I_n$ based on Theorem~\ref{thm:al}, then $I$ is a latent set of the relationships between $I_1, \cdots, I_n$. Now, as $P$ is a primitive, $O$ is a latent set of $I$, and by Theorem~\ref{thm:latent_composition} $O$ is a latent set of the relationships between $I_1, \cdots, I_n$.
\end{proof}

\section{Case Studies}\label{sec:cases}
In this section we will present some architectures built using a primitive, discriminatory pyramids and associative layers. The goal of this section is to present some examples of the versatility of our framework under different scenarios, and what results we expect to obtain, to show the potential of using the \framework. However, proving these are the actual results we will obtain will be matter of future work.

\subsection{The Unsupervised Scenario}
Given an unsupervised scenario, where we have a bunch of data and no label to associate to such data, we can use a discriminatory pyramid to build archetypes of the data. In that sense, the discriminatory pyramid would build archetypes of the possible classes present in the data, based on their relatedness. This is derived from the fact that the output of a discriminatory pyramid is a latent set of the input set, that is, the data. Thus, in some sense, that latent set has merged related samples into classes or clusters.

Here and in the rest of this section, by relatedness we want to mean the function that the primitive uses to decide which samples correspond to the same archetype and which samples do not. For example, an easy function will be the distance function, and then the relatedness will be their closeness between the samples in their representation space. If, for example, the function would be the arrival time, then the relatedness will be the closeness in time of the samples.

A more elaborated unsupervised scenario is where we have data that is not labelled, but that we know there is a relationship between them and we can split them into related values. For example, when we have tabular data where each column is a different feature. In this case, we can build a discriminatory pyramid for each feature, and then add a final associative layer that receives as input each of the outputs of the discriminatory pyramids. With this architecture, each discriminatory pyramid will build archetypes of their feature, and the associative layer will build archetypes of relationships between archetypes of features. To increase more the archetypal power of the framework, we could even add a last primitive that processes the archetypes produced by the associative layer.

In the same fashion than the previous example, this architecture has the potential to, in a totally unsupervised manner, find classes between the inputs. As each discriminatory pyramid has created a latent set of their inputs, they have build archetypes (or classes) of its inputs. Later, the associative layer takes those archetypes, associate them, and generates a latent set of those associations, thus generating archetypes of the associations between archetypes. These in fact behave like classes of the original inputs.

\subsection{The Supervised Classification Scenario}
If we have a classification scenario, we have the actual labels of the classes we want to learn to identify. Thus, there is a mandatory need for, at least, one associative layer. We can even build a bare associative layer, where the inputs of the layer are the samples in one side and the labels in the other. This architecture will associate labels to samples directly and then build archetypes of those relationships. It is not the ideal architecture, but it could work in some scenarios.

A more elaborated architecture would be to provide the samples to a discriminatory pyramid and the labels to another one (although this last pyramid can be avoided if the labels are too simple, like just numbers). Then, the output of both pyramids is provided to the associative layer, that in this case will build relationships between archetypes of the samples and their corresponding label. This in fact has more potential of finding correct classifications than the previous example, as working with archetypes of the samples is expected to allow for better robustness against outliers and noise.

If we keep rising the complexity of the architecture, we can consider to use a discriminatory pyramid for different features of the sample, thus having as many pyramids as features, and then joining all of them and the one for the labels with an associative layer. In this case, the associations will be not only with the label, but also between the features of the sample. Thus, there will be a more fine-grained associativity between a label and the different features of the samples.

Finally, more in general, we can assume that the sample archetype generation is an unsupervised task, that later is associated with the label for classification purposes. This implies that we will identify the characteristics of our sample before assigning them a label. In this case, the architecture would be one of the architectures of the previous section for the sample, whose output is later provided as one of the inputs of an associative layer that also receives the labels. Thus, we will be generating associations between a more refined archetype of the sample and its corresponding label.

This last proposal could imply having an associative layer at the end of the sample architecture, whose outputs are fed to another associative layer. This in fact could be very useful to find associations between a sample features in a first step, then make archetypes of those associations, and finally make associations between those archetypes and the label.

\subsection{The Supervised Regression Scenario}\label{sec:regression}
In a regression scenario, we have a sample and its associated regression value. In this case, we can build similar architectures to the scenario from the previous section, with a particularity: we always need to replace the label (or the discriminatory pyramid of the label) for a discriminatory pyramid for the regression values. This would provide the associative layer with archetypes of the possible regression values, based on their relatedness. This adds some complexity to the framework, as not any primitive that fulfil our requirements can perform a regression effectively. In fact, such primitive will, most probably, have to base its relatedness in some kind of similarity function based on the closeness between the numbers, although properly proving this would be matter of future work.

\subsection{The State-Action Scenario}
Finally, when we have a state-action scenario, we actually have a set of input states and a set of output actions, very similar to the supervised classification scenario. However, a key point here is that we do not have any more a simple label, but instead a complex set of actions. Thus, in this scenario we will have an associative layer at the top that will associate states to actions, but ideally its inputs will be archetypes of states and archetypes of actions. To that end, we suggest to consider the generation of both state and action archetypes as unsupervised scenarios, where we build the best architecture in order to produce the more fine-grained archetypes. Then, the last associative layer will relate those state archetypes to those action archetypes, and it will build archetypes of state-action pairs, thus building a policy. Finally, if we add a primitive over the top associative layer, we will be building archetypes of policies, something we expect will resemble an episodic memory.

\begin{figure}[t]
    \centering
    \includegraphics[width=0.9\columnwidth]{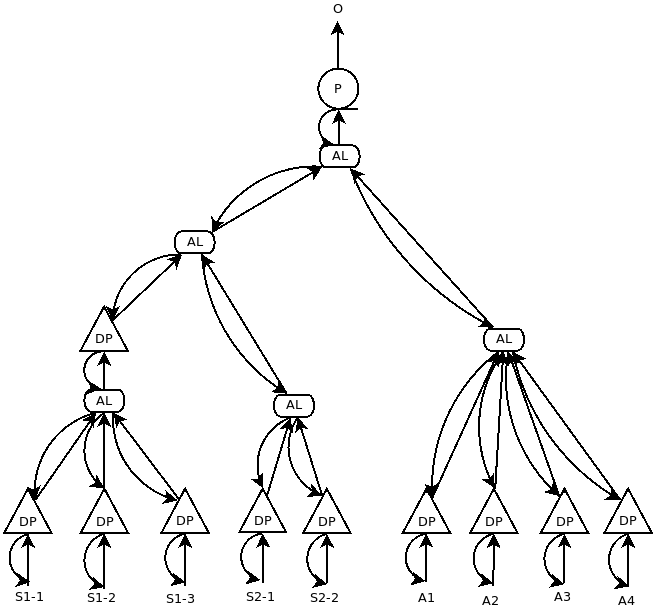}
    \caption{An example of complex architecture.}
    \label{fig:complex}
\end{figure}
An example of this kind of complex architectures is displayed in Figure~\ref{fig:complex}. In that example, we have a left side state analysis architecture, with six lower level discriminatory pyramids, two middle level associative layers, with a discriminatory pyramid on top of one of them, and then a final associative layer to produce the last archetypes of the state. The right side analyses actions, and will either accept inputs for learning, or produce projections when in evaluation mode. This side is composed of four lower level discriminatory pyramids and an associative layer to build archetypes of actions. Finally, we have a top level associative layer that associates state archetypes to action archetypes and builds policies, and a primitive on top to build archetypes of policies.

\subsection{An Illustrative Example}\label{sec:examp}
\begin{figure}[t]
    \centering
    \includegraphics[width=0.9\columnwidth]{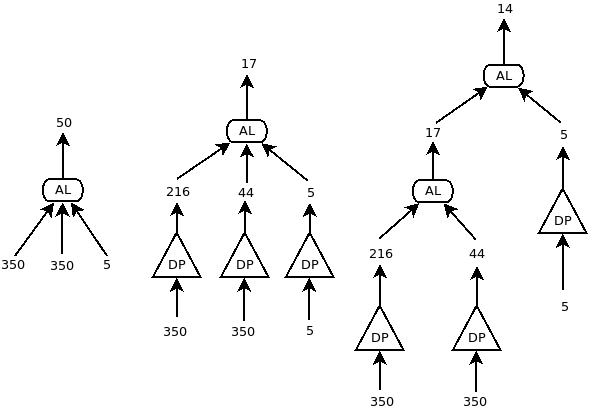}
    \caption{Three architectures for the same dataset. The numbers are the sizes of the corresponding input/output sets.}
    \label{fig:illustrative}
\end{figure}

Finally, as an illustrative example, we will show preliminary results obtained with a primitive implementation that does not fulfil all the requisites presented in this paper, but that has already achieve state-of-the-art results~\citep{iarga25}. To be precise, this primitive has a complex training phase that makes it difficult to fulfil all the requirements during training. However, it gets close enough to fulfil them and it fulfils them once trained. Thus, we consider it valid to be used as an illustrative example. In this example we take a custom dataset that associates digits from the MNIST dataset~\citep{lbbh98}, to hands raising as many fingers as the number, to the corresponding numerical label. It has $350$ samples build with $350$ different MNIST numbers, $350$ different hand samples and $5$ different labels (for numbers from $1$ to $5$). We built three different architectures with our primitive, that showcase the different reduction capabilities of having multiple levels. The first structure has only an associative layer, the second one has an associative layer and a discriminatory pyramid for each modality, and the last architecture has a discriminatory pyramid for each modality, an associative layer that associates MNIST digits to hands, and another associative layer to associate those associations to the label. These structures and their corresponding input and output set sizes are displayed in Figure~\ref{fig:illustrative}.

Due to the characteristics of our primitive, the size of the output set is equivalent to the number of internal representations of the previous primitive. Thus, we can observe how the number of representations is being reduced the further we go in the hierarchy. A one level hierarchy with only one associative layer is able to reduce the number of input samples into a seventh of its size, but it stills obtains a lot of representations. A two level hierarchy that includes also discriminatory pyramids is able to further reduce the number of representations to only $17$ from the initial $350$, what shows the importance of including this kind of structure into an architecture. Finally, a three level hierarchy that includes another associative layer, this time between the discriminatory pyramids of the MNIST digits and the hands and the top associative layer, is able to further reduce the number of representations to only $14$, although the difference with the previous architecture is minimal. Here it is important to remark that the $17$ representations from the second level are not related and are totally different than the $17$ representations from the two level hierarchy.

\section{Limitations}\label{sec:disc}
In this section we want to discuss the limitations of our framework. Specifically, we want to discuss about the fact that our framework is only a theory without empirical validation. In that regard, we have to concede that, even although we proved our claims with mathematical proofs and we have shown some illustrative examples, it is true that some theories can not be implemented. Thus, an empirical validation of our claims is mandatory once we are able to build a primitive that fulfils the requirements presented. However, in this paper we have not included such validation due to the difficulties of building such primitive. We actually have preliminary experiments with primitives that are close to the one required here (as the illustrative example of Section~\ref{sec:examp} presents), and the results are promising and in line with the theory explained here. However, as these experiments are preliminary, with primitives that do not fully fulfil the requirements presented here, we decided not to include them in the form of an empirical evaluation.

\section{Conclusions}\label{sec:conc}
The Artificial Intelligence field lacks a firm proposal of a framework able to build hierarchies of constructive abstractions through archetypes. This is fundamental to develop algorithms and methodologies able to function in a similar fashion to the human brain. Up to date, there have been proposals of building abstractions and proposals of building hierarchies, but no proposal has been made that builds both in a constructive manner and using archetypes, and in this paper we presented the first one.

In this paper, we have presented a framework to generate hierarchies of constructive archetypes based on the assumption of the existence of a primitive with a series of characteristics. Furthermore, we have proven the soundness of our approach with a series of mathematically proven definitions, theorems, lemmas and corollaries. Additionally, we have proved also some properties of our approach, as well as presented some possible alternatives based on the characteristics of the primitive. Finally, we have discussed some limitations of our framework and we have presented some potential architectures based on the use case.

For future work, we would like to develop a primitive that fulfils the requirements we have set up here, and test its performance against traditional machine learning approaches. We would also like to explore what additional properties and potentials can arise from adding new requirements to the primitive, like the primitive for regression scenarios we talked about before in Section~\ref{sec:regression}. We would like to explore methodologies to define the architecture that would work better for a given problem too, even considering methods to automatically build those architectures in a generative way. Finally, we would like to explore the implementation of the architectures presented in Section~\ref{sec:cases}.

\section*{Acknowledgements}
We want to thank Roger Aylagas-Torres, Daniel Pinyol and Pere Mayol for our insightful discussions about the topic.

\bibliographystyle{elsarticle-num}
\bibliography{biblio}

\end{document}